\documentclass[10pt,twocolumn,letterpaper]{article}

\usepackage[pagenumbers]{cvpr}

\usepackage{graphicx}
\usepackage{amsmath}
\usepackage{amssymb}
\usepackage{booktabs}
\usepackage{times}
\usepackage{epsfig}
\usepackage{listings}
\usepackage{multirow}
\usepackage[table,xcdraw]{xcolor}
\usepackage{verbatim}
\usepackage{color, colortbl}
\usepackage{bm}
\usepackage{enumitem}
\usepackage{pifont}
\usepackage[accsupp]{axessibility}
\usepackage{array}
\usepackage{float}
\usepackage[pagebackref,breaklinks,colorlinks]{hyperref}
\usepackage[font=small,labelfont=small]{caption}
\usepackage{algorithm}
\usepackage{algpseudocode}

\definecolor{Gray}{gray}{0.9}
\newcommand{\cmark}{\ding{51}}%
\newcommand{\xmark}{\ding{55}}%

\newcommand{\bev}{\textit{BEV}\xspace}

\newcommand{\LD}{$lane$-$divider$\xspace}
\newcommand{\PC}{$ped$-$crossing$\xspace}
\newcommand{\RB}{$road$-$boundary$\xspace}

\newcommand{\bluecell}[1]{\cellcolor[HTML]{DAE8FC}{#1}}

\newcommand{\greencell}[1]{\cellcolor[HTML]{E0EEE0}{#1}}

\newcommand{\model}{MachMap\xspace}
\newcommand{\hdmap}{\textit{HD-map}\xspace}

\def\confName{CVPR}
\def\confYear{2023}

\begin{document}

\title{MachMap: End-to-End Vectorized Solution for Compact HD-Map Construction}
\author{
	Limeng Qiao$^{1, \star, }$\textsuperscript{\ding{41}} \quad Yongchao Zheng$^{2, \star, \dagger}$ \quad Peng Zhang$^{2, \star, \dagger}$ \quad Wenjie Ding$^{1, \star}$ \\
	Xi Qiu$^{1, }$\textsuperscript{\ding{41}} \quad Xing Wei$^2$ \quad Chi Zhang$^1$ \\
	$^1$Mach Drive \quad $^2$Xi'an Jiaotong University \\
	{\tt\small \{limeng.qiao, wenjie.ding, xi.qiu, chi.zhang\}@mach-drive.com} \\
	{\tt\small \{zyc573823770, zp5070\}@stu.xjtu.edu.cn} \quad {\tt\small weixing@mail.xjtu.edu.cn}
}
\maketitle
	
\begin{abstract}
	This report introduces the $1^{st}$ place winning solution for the Autonomous Driving Challenge $2023$ - Online \hdmap Construction$^\flat$. 
	By delving into the vectorization pipeline, we elaborate an effective architecture,  termed as \model, which formulates the task of \hdmap construction as the point detection paradigm in the bird-eye-view space with an end-to-end manner.
	Firstly, we introduce a novel map-compaction scheme into our framework, leading to reducing the number of vectorized points by $93$\% without any expression performance degradation.
	Build upon the above process, we then follow the general query-based paradigm and propose a strong baseline with integrating  a powerful CNN-based backbone like InternImage, a temporal-based instance decoder and a well-designed point-mask coupling head.
	Additionally, an extra optional ensemble stage is utilized to refine model predictions for better performance.
	Our MachMap-tiny with IN-$1$K initialization achieves a mAP of $79.1$ on the Argoverse$2$ benchmark and the further improved MachMap-huge reaches the best mAP of $83.5$, outperforming all the other online \hdmap construction approaches on the final leaderboard with a distinct performance margin ($ > 9.8$ mAP at least).
\end{abstract}

\footnotetext{$^\dagger$ Work done during an internship at Mach Drive}
\footnotetext{$^\star$ Equal Contribution Authors \quad \text{\ding{41}} ~Corresponding Authors}
\footnotetext{$^\flat$ \scriptsize\url{https://opendrivelab.com/AD23Challenge.html\#Track2}}
	
\section{Introduction}
\label{sec:intro}

\begin{figure}[htb]
	\begin{center}
		\includegraphics[width=1.0\linewidth]{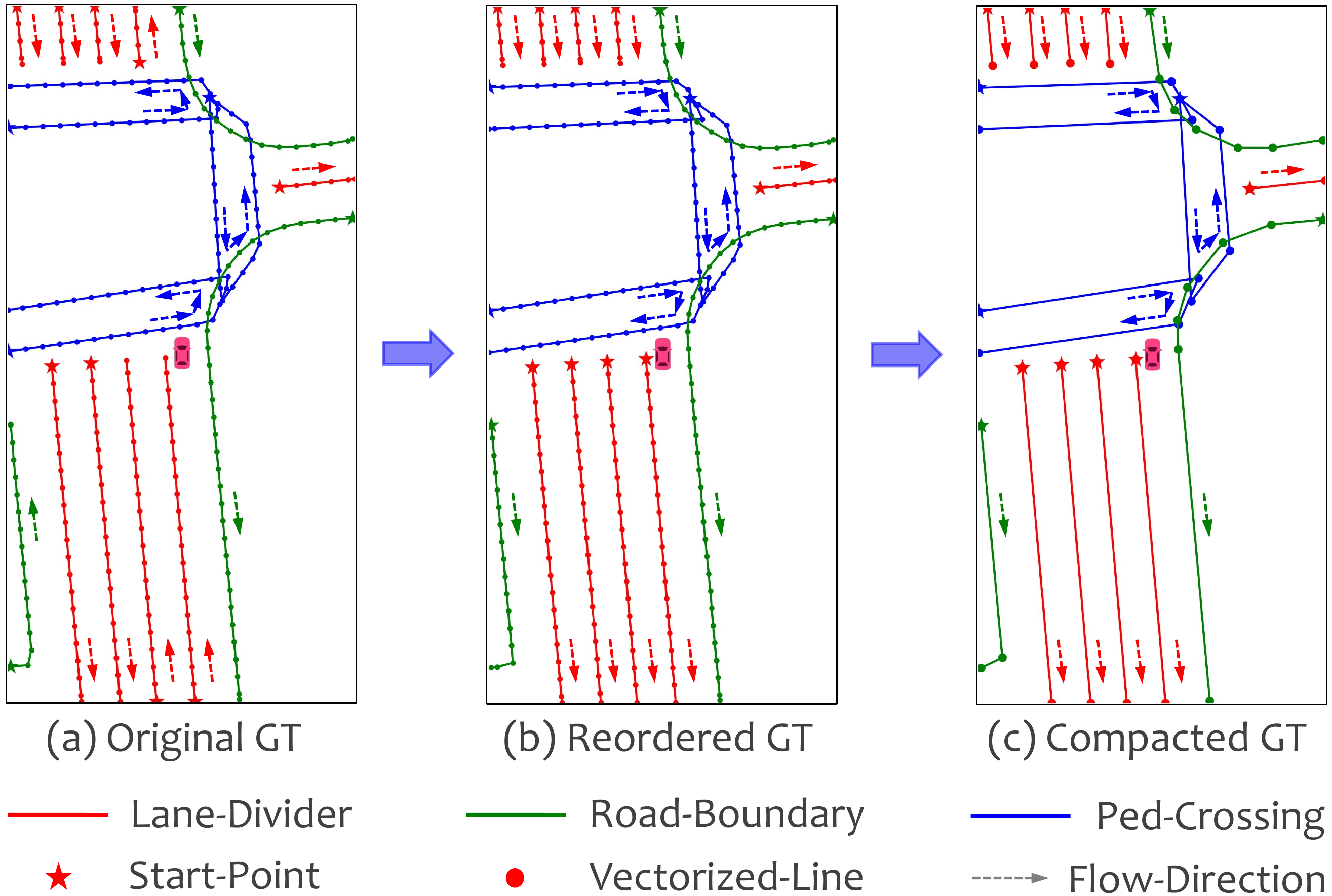}
	\end{center}
	\vspace{-0.4cm}
	\caption{
		\textbf{The illustration of our \hdmap processing principles}.
		 \textit{(a)} the original ground truth given in challenge.
		 \textit{(b)} reorder polylines to keep the \textit{inter-element direction consistency}.
		 \textit{(c)} remove redundancy to keep the \textit{intra-element sequence compactness}.
	}
	\vspace{-0.4cm}
	\label{fig:motivation}
\end{figure} 

As one of the fundamental modules in the autonomous-driving, high-definition map (\hdmap) provides centimeter level environment information for ego-vehicle navigation, including detailed geometric-topology relationships and semantic map categories, \eg \PC, \LD and \RB. Recently, with the development of deep neural network, online construction of local \hdmap from onboard sensors (cameras) has gradually become a more advantageous and potential solution.

\begin{figure*}[htb]
	\begin{center}
		\includegraphics[width=1.0\linewidth]{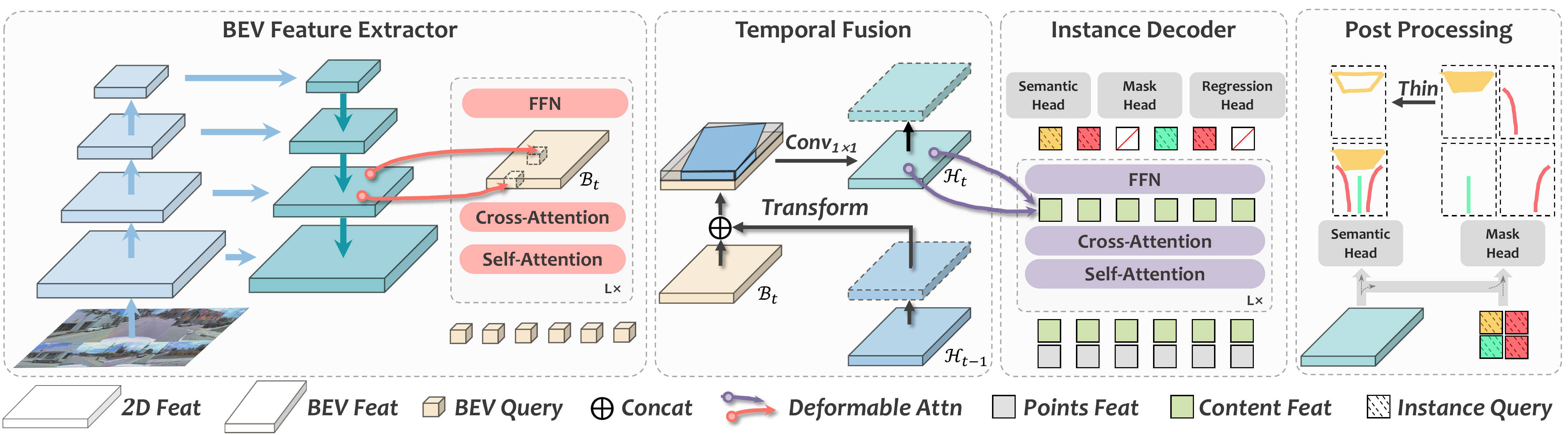}
	\end{center}
	\vspace{-0.5cm}
	\caption{
		\textbf{The architecture of our proposed \textbf{\textit{MachMap}}.}
		Given surrounding images, we generate 2D features from each of views through image backbone and neck.
		Then the deformable attention are used to aggregate the 3D feature among different views and average it along z-axis.
		The temporal fusion module fuses the new BEV feature with the one of hidden state of BEV feature, based on which the hidden state is updated.
		Finally, we conduct instance decoder which utilize instance-level deformable attention to refine content and points features and format the final results.
		It is worth noting that the results of \PC and \LD are thinned from the mask.
	}
	\vspace{-0.5cm}
	\label{fig:framework}
\end{figure*} 

The Online \hdmap Construction Track aims to dynamically construct local \hdmap from onboard surrounding camera images. In this task, a local \hdmap ground truth in \textit{Fig.\ref{fig:motivation} (a)} is described by a set of map elements with three semantic categories and each element is designed to a polyline, which consists of a set of ordered points, to deal with complicated and even irregular road structures. Our method mainly focuses on three aspects to handle the competition,

\noindent
\textbf{\textit{(1) map modeling principles.}} 
We propose the principles of \textit{inter-element direction consistency} and \textit{intra-element sequence compactness} to reduce the intrinsic redundancy of polyline-based map modeling. Concretely, without losing any expression performance, the flow directions of point sequences between different elements should be as consistent as possible, and the point sequences within the same map element should be reserved with as few points as possible.

\noindent
\textbf{\textit{(2) temporal-fusion instance decoder.}}
Based on the multi-cameras features from image backbone, we then employ a temporal-fusion based bird-eye-view (\bev) feature decoder for view-transformation and a bottom-up point-wise instance decoder to extract point descriptor.

\noindent
\textbf{\textit{(3) point-mask coupling head.}}
Considering that different map elements have distinct shape priors, \eg \LD is usually polyline and \PC is convex polygon, we equip each semantic map category with both segmentation and detection heads under the MaskDINO \cite{maskdino} framework, which greatly improves the flexibility and scalability of our model. Furthermore, the above multi-task training strategy also accelerates the model convergence performance.

Inspired by the above motivations, we propose an end-to-end vectorized \hdmap construction architecture, named as \textbf{\textit{\model}}. The entire framework is illustrated in \textit{Fig.\ref{fig:framework}} and all technical details are presented in the next section.
	
\section{Method}
\label{sec:method}

This section introduces the details of our winning method.
We first present the map compaction pipeline, which significantly reduces the difficulty of model training and makes the inference results more compact and efficient.
Next the design scheme of each module is presented, and some task-specific improvements are integrated into some off-the-shelf methods.
Lastly, we introduce our novel ensemble ideas, which can further enhance our approach.

\subsection{Map Compaction Pipeline}
Different from rasterized scheme, vectorized \textit{HD-maps} in the given annotations explicitly express the spatial relation between map elements and instance information in their respective categories.
Following the newly proposed map modeling principles, we compact the original evenly sampled map representation in two steps, namely orientation rearrangement and redundancy removal.

\noindent
\textbf{\textit{(1) inter-element direction consistency.}}
The directions of elements in original map annotations are in a state of chaos, such as \textit{lane-dividers} of moving forward from front-to-back or back-to-front as shown in \textit{Fig.\ref{fig:motivation} (a)}. We noticed that the inconsistency of directions can negatively affect the training of the model. To reduce the discreteness of map organization, we follow a certain strategy to make the orientation of map elements as orderly as possible, and guarantee that this process does not lose any details of the map. 
Specifically, under the principle of conforming to the observation order of human eyes, a simple and intuitive strategy is to reorganize all polylines according to the rules \textit{from-front-to-back} and \textit{from-left-to-right} in bird-eye-view space.

\noindent
\textbf{\textit{(2) intra-element sequence compactness.}} 
Vectorized maps with evenly-distributed points have redundant semantic information, while compacted-points representation is sparse, which is more suitable for expression and storage of maps.  To this end, we extract  keypoints for all elements to supervise model training. Concretely, we adopt Douglas-Peucker algorithm~\cite{visvalingam1990douglas} and Visvalingam algorithm~\cite{visvalingam1995simplification} to condense a polyline composed of line segments to a similar polyline with fewer points. 
For these methods, points are removed in order of least to most importance, with importance related to the distance and triangular area respectively.

\subsection{MachMap Architecture}

We follow the general query-based design paradigm~\cite{dab-detr, bemapnet}, as illustrated in the \textit{Fig.\ref{fig:framework}}, where the overall structure can be roughly divided into three parts: \textit{BEV feature extractor}, \textit{temporal-fusion instance decoder}, and \textit{point-mask coupling head}.
Afterwards, we introduce each module sequentially according to the flow of information.

\noindent
{\textbf{Backbone.}}
Giving a list of $2$D images $\mathcal{I} \in \mathcal{R}^{N \times 3 \times H \times W}$,
extracting unified textures representation within images is a top-priority task.
With regards to this, we utilize a shared InternImage~\cite{intern} as strong backbone to extract image features, which employs deformable convolutions~\cite{dai2017deformable} as its core operator and has been meticulously designed.
During the downsampling process, a series of feature maps in varying scales are generated and then aggregated by the Bi-directional Feature Pyramid Network, \ie BiFPN~\cite{bifpn}.

\noindent
{\textbf{Multi-view Encoder.}}
Since the map vectors we ultimately need to predict lay in $3$D space, it is necessary to elevate surrounding features from camera-view to $3$D ego-view space.
Rather than direct transformation to $3$D-view, we predefine a set of reference points and arrange them in a \bev raster.
After that, we employ the camera intrinsics and extrinsics to project them onto several images and then  aggregate the surrounding features. By averaging on the \textit{z}-axis, we obtain the final bird-eye-view features $\mathcal{B} \in \mathcal{R}^{H_{B} \times W_{B} \times C}$.

\noindent
{\textbf{Temporal Fusion Module.}}
The provided dataset is collected and organized chronologically, with precise poses for each sample.
This makes it possible to align current features with previous ones by poses, resulting in a larger real-world perception range beyond the current position.
We follow the long-term fusion strategy proposed in VideoBEV~\cite{videobev}, which affines the previous hidden state $\mathcal{H}_{t-1}$ of \bev feature into the current one $\mathcal{B}_{t-1}$ using vehicle ego pose.
The latter is concatenated with the current \bev feature $\mathcal{B}_{t}$ in the channel dimension and fused by a $1 \times 1$ convolutional layer as,
\begin{align}
	\label{eq:curve}
	\mathcal{B}_{t-1} &= \texttt{Affine}_{\textit{pre} \rightarrow \textit{cur}}(\mathcal{H}_{t-1}) \\
	\mathcal{H}_{t} &= \texttt{Conv}_{1 \times 1}(\mathcal{B}_{t-1} \oplus \mathcal{B}_{t})
\end{align}
where $\oplus$ denotes the concatenation operator. 
The fused features are cached as the next hidden state and used as input for subsequent instance decoder.
In practice, since the timestamp offset between adjacent frame is too small, we group the timestamps at specific intervals to expand the performance gain brought by this temporal-fusion module.

\noindent
{\textbf{Instance Decoder.}}
To benefit from multi-task loss, we opt for the MaskDINO ~\cite{maskdino} framework, which conducts object detection and segmentation tasks simultaneously. Each query consists of content and position vectors, with the former is utilized to generate instance masks, while the latter undergoes iterative updates to yield normalized coordinates directly. Yet, due to the hierarchical relationship between map elements and their corresponding points sets, we adopt the query design paradigm in MapTR~\cite{liao2022maptr} for better adaptation to map element modeling. This implies that the query is point-wise, and a set of which can be aggregated to form a single instance and obtain its corresponding instance mask.

\noindent
{\textbf{Output Head.}}
Using only coordinates from point regression has some drawbacks.
Firstly, there is a keypoint mismatching issue, where a well predicted instance may occur a mismatched point which belongs to other instance, as a result, a single bad apple spoils the whole bunch. 
Secondly, for \PC, there exists a strong geometric prior, which is difficult to depict through vectors. However, masks not only can effectively constrain the geometry shape of instances, but it also impose a significant penalty on mismatched points during training. Empirically, we obtain \PC and \LD through post-processing of instance masks, while point regression is employed only for \RB. As the common practice, we adopt cross-entropy and dice loss~\cite{dice} for masks and L$1$ loss for point regression. In addition, we also add semantic loss to the \bev features as auxiliary supervision, and our final loss as,
\vspace{-0.55cm}
\begin{align}
	\label{eq:loss}
	\mathcal{L}=\lambda_\textit{cls}\mathcal{L}_\textit{cls}+\lambda_\textit{pts}\mathcal{L}_\textit{pts}+\lambda_\textit{mask}\mathcal{L}_\textit{mask}+\lambda_\textit{sem}\mathcal{L}_\textit{sem}
	\vspace{-0.25cm}
\end{align}
where $\lambda_{\star}$ is the balance weight for different losses.

\subsection{Ensemble Strategy}
The predicted map vectors of our model are represented in normalized coordinates, which are then rescaled to the actual range $60 \times 30 m$ in the ego coordinate system during the post-processing stage. Yet the actual visible content from images greatly exceeds this range, which often leads to ambiguities in the existence of certain elements at the border position of exact map region that may be ignored by a single model. Accordingly, the use of ensemble techniques can mitigate prediction variability and curb overfitting by summarizing multiple models together.

By utilizing chamfer distance as a metric for measuring the similarity between instances, we present the ensemble algorithm in the Algorithm \ref{alg:cap}.
Given a base set and a list of proposals, which are derived from multiple other predictions and sorted by confidence in descending order, we can compare each proposal with the base set one by one.
If their similarity is low, we can consider them as missed true positives and add them to the base set.
In addition to multi-model ensemble, we also conduct multi-frame ensemble. Despite the utilization of temporal fusion module, some instances are still absent, which were accurately recalled in previous frames. This inspires us to compensate some erratic predictions by ensemble with predictions from previous frames. It's worth noting that the integration of multi-frame and multi-model can share the same algorithm, with only modifying the source of candidate proposal list.

\renewcommand{\algorithmicrequire}{\textbf{Input:}}
\renewcommand{\algorithmicensure}{\textbf{Output:}}
\begin{algorithm}
	\footnotesize
	\caption{MachMap Ensemble Algorithm}\label{alg:cap}
	\begin{algorithmic}[1]
		\Require Base-list $B$, Proposal-list $P$ and score-list $S$, CD-threshold $T$
		\Ensure Added proposal list and score $A,AS$
		\State $P,S \gets$\texttt{SortProposalByScore}$(P,S)$
		\State $A \gets [], AS \gets []$
		\While{$P$\texttt{.length} $ \neq 0$}
		\State $Flag \gets False$
		\State $Head \gets P$\texttt{.pop} $,HeadScore \gets S$\texttt{.pop}
		\For{$Base$ \texttt{in} $B$}
		\State $Sim \gets$ \texttt{ChamferDistance}$(Head,Base)$
		\If{$Sim < T$}
		\State $Flag \gets True$
		\State \textbf{break}
		\EndIf
		\EndFor
		\If {$not \ Flag$}
		\State $B$\texttt{.append}$(Head)$
		\State $A$\texttt{.append}$(Head)$
		\State $AS$\texttt{.append}$(HeadScore \cdot \sigma)$ \Comment{$\sigma$ is a score decay factor}
		\EndIf
		\EndWhile
	\end{algorithmic}
\end{algorithm}

\begin{table*}[h]
	\begin{center}
		\resizebox{0.92\textwidth}{!}{
			\begin{tabular}{c|cccc|cccc}
				\hline
				\rowcolor{Gray}
				Category & \# images & \# instances & \# points (raw) & \# points (compacted) & AP$_{0.2m}$ & AP$_{0.3m}$ & AP$_{0.4m}$ & AP$_{0.5m}$  \\
				\toprule
				\PC & 19523 & 55686 & 3593548 & 252219~\textbf{\scriptsize{\color{blue}($\downarrow93.0\%$)}} & 98.33 & 99.46 & 99.92 & 100.00 \\
				\LD & 26222 & 133186 & 7426425 & 335534~\textbf{\scriptsize{\color{blue}($\downarrow95.5\%$)}}  & 99.91 & 99.98 & 99.99 & 100.00 \\
				\RB & 27283 & 84384  & 7018193 & 469533~\textbf{\scriptsize{\color{blue}($\downarrow93.3\%$)}}  & 97.38 & 99.70 & 99.92 & 100.00 \\
				\bottomrule
			\end{tabular}
		}
	\end{center}
	\vspace*{-0.5cm}
	\caption{
		\textbf{The effectiveness and correctness verification of map compaction principles.}
		All statistical numbers are collected on both training and validation sets. Note that \# means 'the number of' and the blue color means the proportion of point reduction. AP$_\tau$ indicates that the average precision between before and after the compaction, where a prediction as true-positive only if the distance is less than $\tau$.
	}
	\vspace*{-0.2cm}
	\label{tab:compaction}
\end{table*}

\begin{table*}[h]
	\begin{center}
		\resizebox{0.92\textwidth}{!}{
			\begin{tabular}{c|ccccc|ccc|c}
				\hline
				\rowcolor{Gray}
				ID & Data & Backbone & PreTrain & \# Epochs & w/o \textit{Opt.} & AP$_\textit{{crossing}}$ & AP$_\textit{divider}$ & AP$_\textit{{boundary}}$ & mAP  \\
				\toprule
				$1$ & \textit{train} & \textit{tiny} & ADE$20$K & 6 & \xmark & 61.01 & 65.87 & 65.70 & 64.19 \\
				$2$ & \textit{train} & \textit{tiny} & ADE$20$K & 72 & \xmark & 76.75 & 73.51 & 74.68 & 74.98 \\
				$\square$ & \textit{train+val} & \textit{tiny} & ADE$20$K & 72 & \xmark & 78.34 & 74.74 & 76.02 & 76.37 \\
				$3$ & \textit{train+val} & \textit{tiny} & $\textit{from - }\square$ & $\rightsquigarrow$ 6 & \cmark & 84.82 & 79.66 & 80.63 & 81.70 \\
				$\heartsuit$ & \textit{train+val} & \textit{huge} & ADE$20$K & 36 & \xmark & 81.45 & 75.34 & 77.14 & 77.98 \\
				\bluecell{$4$} & \bluecell{\textit{train+val}} & \bluecell{\textit{huge}} & \bluecell{\textit{from - }$\heartsuit$} & \bluecell{$\rightsquigarrow$12} & \bluecell{\cmark} & \bluecell{\textbf{86.66}} & \bluecell{\textbf{81.54}} & \bluecell{\textbf{82.29}} & \bluecell{\textbf{83.50}} \\
				\midrule
				$\triangle$ & \textit{train+val} & \textit{tiny} & IN-$1$K & 72 & \xmark & 76.46 & 72.32 & 75.91 & 74.90 \\
				\greencell{$5$} & \greencell{\textit{train+val}} & \greencell{\textit{tiny}} & \greencell{\textit{from - }$\triangle$} & \greencell{$\rightsquigarrow$ 6} & \greencell{\cmark} & \greencell{82.01} & \greencell{76.23} & \greencell{79.10} & \greencell{79.11} \\
				\bottomrule
			\end{tabular}
		}
	\end{center}
	\vspace*{-0.5cm}
	\caption{
		\textbf{The performance of different \textit{\model} milestone models} under thresholds of $[0.5, 1.0, 1.5]m$.
		we employ InternImage \cite{intern} as backbone and '\textit{tiny}/\textit{huge}' means its scale.
		The '\textit{from - }$\square$/$\heartsuit$/$\triangle$' means loading corresponding checkpoint and $\rightsquigarrow$ is more epochs finetuning. Note the weights of  ADE$20$K/IN-$1$K are public. 
		The term \textit{Opt.} means our improving techniques, \eg \textit{ema}, \textit{ida}, \textit{temporal} and \textit{ensemble}.
	}
	\vspace*{-0.41cm}
	\label{tab:main-result}
\end{table*}

\section{Experiments}
\label{sec:experiments}

\subsection{Existing Benchmarks}
The Argoverse$2$ ~\cite{Argoverse} contains $700$, $150$ and $150$ video clips in the training, validation, and testing sets respectively.
Each sequence has $6$-DOF map-aligned pose and seven ring views with the image resolution of $2048 \times 1550$ or $1550 \times 2048$ pixels. The given data from challenge is a subset of Argoverse$2$.
We utilize all frames from the challenge training set to verify the effect of different ablations but finally all frames from training and validation sets are used to reach better performance. 
We focus on three categories, \ie \LD, \PC and \RB.

\subsection{Implementation Details}

\noindent
{\textbf{Training Setup.}}
We adopt common data augmentation, \eg random scaling, cropping, and flipping.
At the same time, an \textit{IDA}~\cite{bevdet} matrix is updated to record view transformation to maintain spatial consistency.
Then the final input shape is fixed at $896 \times 768$, as this aspect ratio is close to the front view, \ie $2048 \times 1550$, which contains the most abundant visual map information.
For \bev features, the default spatial shape of \bev queries is $64 \times 32$, which corresponds to the perception ranges in lidar coordinate system are $[-30, 30]m$ for the \textit{Y}-axis and $[-15, 15]m$ for the \textit{X}-axis.
Note all map masks are interpolated to $400 \times 200$ to ensure that distinct elements can be easily distinguished without occupying too much memory.
As for the hyperparameters of loss function, we set $\lambda_\textit{cls}$, $\lambda_\textit{pts}$, $\lambda_\textit{mask}$, $\lambda_\textit{sem}$ to $2$,  $20$,  $1$, and $3$ respectively.

\begin{table}[!htb]
	\begin{center}
		\resizebox{1.0\linewidth}{!}{
			\begin{tabular}{cc|ccc|c}
				\hline
				\rowcolor{Gray}
				Rank & Team & AP$_\textit{{crossing}}$ & AP$_\textit{divider}$ & AP$_\textit{{boundary}}$ & mAP  \\
				\toprule
				\bluecell{$1$} & \bluecell{Mach (ours)} & \bluecell{\textbf{86.66}}  & \bluecell{\textbf{81.54}} & \bluecell{\textbf{82.29}}  & \bluecell{\textbf{83.50}} \\
				$2$ & MapNeXt & 68.94  & 76.66 & 75.34 & 73.65 \\
				$3$ & SCR & 70.37 & 75.08 & 74.73 & 73.39 \\
				$4$ & LTS & 72.67 & 73.20 & 71.80 & 72.56 \\
				$5$ & USTC-VGG & 69.05  & 73.24 & 70.76 & 71.02  \\
				\bottomrule
			\end{tabular}
		}
	\end{center}
	\vspace*{-0.5cm}
	\caption{Top 5 entries on the test leaderboard of challenge.}
	\vspace*{-0.41cm}
	\label{tab:leaderboard}
\end{table}

\noindent
{\textbf{Training Strategy.}}
We train our model with a total batch of $8$ on $8$ GPUs.
The AdamW~\cite{adamw} optimizer is employed with a weight decay of $5 \times 10^{-2}$ and a learning rate of $3 \times 10^{-4}$.
Our training process consists of two stages: base training and fine-tuning.
Firstly, we initialize the InternImage~\cite{intern} with public pretrained weights \cite{deng2009imagenet, zhou2017scene}  and then train our model for $60$ epochs without any tricks except a multi-step schedule with milestone $[0.7,0.9]$ and $\gamma = \frac{1}{5}$.
Afterward, we apply all proposed improving techniques to fine-tune the model for extra epochs with a learning rate of $1 \times 10^{-4}$.

\subsection{Experimental Results}

\noindent
{\textbf{Table 1.}} 
Our statistical results show the compacted map can reduce more than $93\%$ points without expression performance losing under the threshold of $0.5m$, even it can still maintain more than $97\%$ performance under stricter $0.2m$.

\noindent
{\textbf{Table 2.}}
Comparing the results in \textit{row-}$1\&2$, training more epochs brings a performance gain of more than $10$ points, which shows that accelerating the convergence speed is still a vital future work. 
Compared with \textit{row-}$3\&4,5\&6,7\&8$, using the proposed improving techniques can always bring more than $5$ points of increase. Moreover, even starting with IN-$1$K as pretrained weights, our model still achieves 79.1.

\noindent
{\textbf{Table 3.}}
We succeed the championship with a performance advantage of $9.85$ mAP over the second place, demonstrating the effectiveness of our proposed \textit{\model} method.

\clearpage

{\small
	\bibliographystyle{ieee_fullname}
	\bibliography{egbib}
}
	
\end{document}